\documentclass{article}
\usepackage{ijcai17}
\usepackage{latexsym}
\usepackage{graphicx}
\usepackage{multirow}
\usepackage{times}
\usepackage{url}
\usepackage{subfigure}
\usepackage{amsmath}
\usepackage[small]{caption}

\title{Interactive Attention Networks for Aspect-Level Sentiment Classification}

\author{Dehong Ma$^1$, Sujian Li$^{1,2}$, Xiaodong Zhang$^{1}$, Houfeng Wang$^{1,2}$\\ 
$^1$MOE Key Lab of Computational Linguistics, Peking University, Beijing, 100871, China \\
$^2$Collaborative Innovation Center for Language Ability, Xuzhou, Jiangsu, 221009, China\\
\{madehong, lisujian, zxdcs, wanghf\}@pku.edu.cn}
\begin{document}

\maketitle

\begin{abstract}
Aspect-level sentiment classification aims at identifying the sentiment polarity of specific target in its context.
Previous approaches have realized the importance of targets in sentiment classification and developed various methods with the goal of precisely modeling their contexts via generating  target-specific representations.
However, these studies always ignore the separate modeling of targets.
In this paper, we argue that both targets and contexts deserve special treatment and need to be learned their own representations via interactive learning.
Then, we propose the  interactive attention networks (IAN) to interactively learn attentions in the contexts and targets, and generate the representations for targets and contexts separately.
With this design, the IAN model can well represent a target and its collocative context, which is helpful to sentiment classification.
Experimental results on SemEval 2014 Datasets demonstrate the effectiveness of our model.
\end{abstract}

\section{Introduction}
\label{section:intro}
Aspect-level sentiment classification is a fine-grained task in
sentiment analysis which aims to identify the sentiment polarity of 
targets in their context~\cite{INR-011,liu2012sentiment}. 
For example, 
Given the mentioned targets: \emph{staff}, \emph{pizza} and \emph{beef cubes}, and
their context sentence \emph{ ``a group of friendly staff, the pizza is not bad,
but the beef cubes are not worth the money!''}, 
the sentiment polarity for the three targets, \emph{staff}, \emph{pizza} and \emph{beef cubes}, are positive, neutral and negative respectively.

Aspect-level sentiment classification is a fundamental task in natural language processing and catches many researchers' attention. 
Traditional approaches mainly focus on designing a set of
features such as bag-of-words, sentiment lexicon to train a
classifier (e.g., SVM) for aspect-level sentiment classification
~\cite{jiang2011target,perez2012learning}.
However, feature engineering is labor intensive and almost reaches its performance bottleneck.
With the development of deep learning techniques, some
researchers have designed effective neural networks to
automatically generate useful low-dimensional representations from targets and their contexts and obtain a promising
result on the aspect-level sentiment classification task
~\cite{dong2014adaptive,vo2015target,wang2016attention,tang2016aspect}.

As Jiang et al. \shortcite{jiang2011target} point out that 40\% of sentiment classification errors are caused by not considering
targets in sentiment classification, recent work tends to especially strengthen the effect of targets when modeling the contexts.
Dong et al. \shortcite{dong2014adaptive}  propose an adaptive recursive neural network (RNN) to propagate the sentiments  from context  words to specific targets based on syntactic relations on tweet data. 
Vo and Zhang \shortcite{vo2015target}  separate the whole context into three components, i.e., target, its left context and right context, and then use sentiment lexicon and neural pooling functions to generate the target-dependent features.
Tang et al. \shortcite{tang2015effective} divide the contexts into  left part with target and right part with target and use two long short-term memory (LSTM) models to model the two parts respectively. 
Then the composed target-specific representations from both parts are used for sentiment classification.
Wang et al. \shortcite{wang2016attention}  design aspect embeddings for targets and concatenate  them with word representations to generate the final representations using LSTM networks and attention mechanism.

The  studies above have realized the importance of targets and developed various methods with the goal of precisely modeling contexts via generating  target-specific representations.
However, they all ignore the separate modeling of targets, especially with the aid of contexts.
In our opinion, only the coordination of targets and their contexts can really enhance the performance of sentiment classification.
Let us take ``The picture quality is clear-cut but the battery life is too short'' as an example.
When ``short'' is collocated with ``battery life'', the sentiment tends to be negative.
But  when  ``short'' is used with ``spoon'' in  the context  ``Short fat noodle spoon, relatively deep some curva'',  the sentiment can be neutral.
Then, the next problem is how to simultaneously model targets and contexts precisely.
First,  target and context can  determine representations of each other.
For example, when we see the target  ``picture quality'',  context word ``clear-cut'' is naturally associated with the target.
And it is vice versa - ``picture quality'' is first connected with  ``clear-cut''.
In such cases, we argue that targets and contexts can be modeled separately but learned from their interaction. 
Second, our common sense is that the context is composed of many words. In fact, targets are also not limited to only one word.
No matter targets or contexts, different words may have different contributions to the final representation.
For example, it is easy to know that ``picture'' plays a more important role in the representation of the  target ``picture quality'' which is described by ``clear-cut''.
Thus,  we first propose that  both targets and contexts should be computed their attention weights to capture their important information respectively.

Based on the two points analyzed above,  we propose an interactive attention network~(IAN) model which is based on long-short term memory networks~(LSTM) and attention mechanism. 
IAN utilizes the attention mechanism associated with a target to get important information from the context and  compute context representation for sentiment classification.
Further, IAN makes use of the interactive information from context to supervise the modeling of the target which is helpful to judging sentiment. 
Finally, with both target representation and context representation concatenated, IAN predicts the sentiment polarity for the target within its context.
Experiments on SemEval 2014 demonstrate that  IAN can precisely model both targets and contexts, and achieve the state-of-the-art performance.

\section{Model}
In this section, we first introduce the architecture of interactive attention networks~(IAN) model for aspect-level sentiment classification.
Next, we show the training details of IAN.
The overall architecture of IAN model is shown in Figure \ref{fig:IAN}.
\subsection{Interactive Attention Networks}
The IAN model is composed of two parts which model the target and context interactively. 
With word embeddings as input, we employ LSTM networks to obtain hidden states of words on the word level for a target and its context respectively.
We use the average value of the target's hidden states and the context's hidden states to supervise the generation of attention vectors with which the attention mechanism is adopted to capture the important information in the context and target. With this design, the target and context can influence the generation of their representations interactively.
Finally, target representation and context representation are concatenated as final representation which is fed to a softmax function for aspect-level sentiment classification.
\begin{figure}[htb]
\centering
\includegraphics[width=1.0\columnwidth]{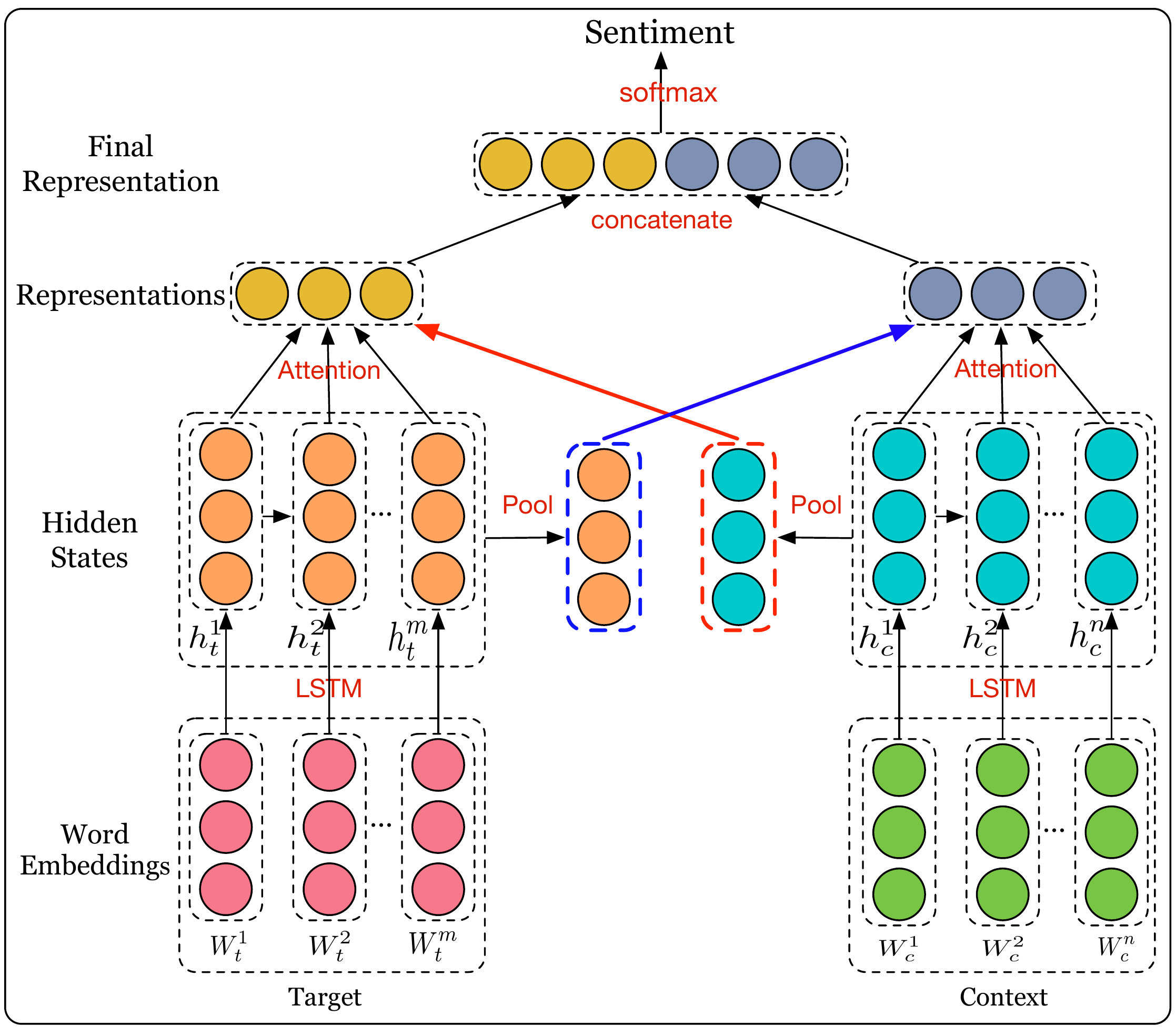}
\caption{The overall architecture of IAN.}
\label{fig:IAN}
\end{figure}

Specifically, let us first formalize the notation. We suppose that a context consists of $n$ words $[w_c^1, w_c^2, ..., w_c^n]$ and a target has  $m$ words $[w_t^1, w_t^2, ..., w_t^m]$. $w$ denotes a specific word.
To represent a word, we embed each word into a low-dimensional  real-value vector, called word embedding~\cite{bengio2003neural}.
Then, we can get  $w^k \in R^d$ from $M^{v\times d}$, where $k$ is the word index in the context or target, $d$ means the embedding dimension and $v$ gives the vocabulary size.  
Word embeddings can be regarded as parameters of neural networks or pre-trained from proper corpus via language model~\cite{collobert2008unified,mikolov2010recurrent,huang2012improving,pennington2014glove}. In our model, we choose the latter strategy.
 
Next, we use the LSTM  networks to learn the hidden word semantics, since words in a sentence have strong dependence on each other, and LSTM is good at learning  long-term dependencies and can avoid gradient vanishing and expansion problems. 
Formally, given the input word embedding $w^k$, previous cell state $c^{k-1}$ and previous hidden state $h^{k-1}$, the current cell state $c^k$ and current hidden state $h^k$ in the LSTM networks are updated as:

\begin{align}
\textbf{i}^k &= \sigma(W_\textbf{i}^w \cdot w^k + W_\textbf{i}^h \cdot h^{k-1} + b_\textbf{i}) &\\
\textbf{f}^k &= \sigma(W_\textbf{f}^w \cdot w^k + W_\textbf{f}^h \cdot h^{k-1} + b_\textbf{f}) &\\
\textbf{o}^k &= \sigma(W_\textbf{o}^w \cdot w^k + W_\textbf{o}^h \cdot h^{k-1} + b_\textbf{o}) &\\
\hat c^k &= tanh(W_c^w \cdot w^k + W_c^h \cdot h^{k-1} + b_\textbf{c}) &\\
c^k &= \textbf{f}^k \odot c^{k-1} + \textbf{i}^k \odot \hat c^k &\\
h^k &= \textbf{o}^k  \odot tanh(c^k)&
\end{align}
where $\textbf{i}$, $\textbf{f}$ and $\textbf{o}$ are input gate, forget gate and output gate respectively, which serve to model the interactions between memory cells and their environments. 
$\sigma$ means a sigmoid function.
$W$and $b$ denote weight matrices and biases respectively.
The symbol $\cdot$ stands for matrix multiplication, and $\odot$ is element-wise multiplication.
Then, we get the hidden states  [$h_c^1, h_c^2, ..., h_c^{n}$] as the final word representations for context.
To better model target's meaning, we also use LSTM networks to obtain the target's hidden states [$h_t^1, h_t^2, ..., h_t^{m}$].

Then, we can get the initial representations of context and target (i.e., $c_{avg}$ and $t_{avg}$ ) by averaging the hidden states.
\begin{align}
c_{avg} = \sum_{i=1}^{n}h_c^i/n &
\end{align}
\begin{align}
t_{avg} = \sum_{i=1}^{m}h_t^i/m &
\end{align}

With  the initial representations of context and target as input,  we adopt the attention mechanism to select important information contributing to judging sentiment polarity. 
As Section \ref{section:intro} stated, we consider the influence on the context from the target and the influence on the target from context,  which can provide more clues to pay attention to those related sentiment features. 

We take a pair of context and target to describe the attention process, as shown in Figure \ref{fig:IAN}.
With context word representations [$h_c^1, h_c^2, ..., h_c^{n}$], the attention mechanism generates the attention vector $\alpha_i$ using target representation $t_{avg}$ for context by:

\begin{equation}
\alpha_i = \frac{exp(\gamma(h^i_c, t_{avg}))}{\sum_{j=1}^{n} exp(\gamma(h^j_c, t_{avg}))}
\end{equation}
where $\gamma$ is a score function that calculates the importance of $h^i_c$ in the context. The score function $\gamma$ is defined as:
\begin{equation}
\gamma(h^i_c, t_{avg}) = tanh(h^i_c \cdot W_a \cdot {t_{avg}}^T + b_a)
\label{eq:gamma}
\end{equation}
where  $W_a$ and $b_a$ are weight matrix and bias respectively, $tanh$ is a non-linear function and ${t_{avg}}^T$ is the transpose of the $t_{avg}$.

Similarly, for the target, we calculate its attention vector $\beta_i$ using context representation $c_{avg}$  by:
\begin{equation}
\beta_i = \frac{exp(\gamma(h^i_t, c_{avg}))}{\sum_{j=1}^{m} exp(\gamma(h^j_t, c_{avg}))}
\end{equation}
where $\gamma$ is the same as in Eq. \ref{eq:gamma}.

After computing the word attention weights, we can get context and target representations $c_r$ and $t_r$ based on the attention vectors $\alpha_i$ and $\beta_i$ by: 
\begin{align}
c_r &= \sum_{i=1}^{n} \alpha_i  h_c^i &\\
t_r &= \sum_{i=1}^{m} \beta_i  h_t^i &
\end{align}

Finally, the target representation $t_r$ and context representation $c_r$ are concatenated as a vector $\textbf{d}$ for a classifier. Here, we use a non-linear layer to project $\textbf{d}$ into the space of the targeted $C$ classes. That is,
\begin{equation}
x = tanh(W_l\cdot \textbf{d}+b_l)
\end{equation}
where $W_l$ and $b_l$ are the weight matrix and bias respectively.
The probability of labeling document with sentiment polarity $i (i \in [1, C])$ is computed by:
\begin{equation}\label{equ:prob}
y_i = \frac{exp(x_i)}{\sum_{i=1}^{C} exp(x_i)}
\end{equation}

The label with the highest probability is set as the final result.

\subsection{Model Training}
In IAN, we need to optimize all the parameters notated as $\Theta$ which are from LSTM networks: [$W_\textbf{i}^w$, $W_\textbf{f}^w$, $W_\textbf{o}^w$, $W_\textbf{c}^w$, $W_\textbf{i}^h$, $W_\textbf{f}^h$, $W_\textbf{o}^h$, $W_\textbf{c}^h$, $b_\textbf{i}$, $b_\textbf{f}$, $b_\textbf{o}$, $b_\textbf{c}$], the attention layers: [$W_a$, $b_a$], the softmax layer: [$W_l$, $b_l$] and the word embeddings.
Cross entropy with $L_2$ regularization is used as the loss function, which is defined as:
\begin{equation}
J = -\sum_{i=1}^{C}g_{i}\log(y_i) + \lambda_r(\sum_{\theta \in \Theta }\theta^2)
\end{equation}
where $g_i \in R^{C}$ denotes the ground truth, represented by one-hot vector; $y_i \in R^{C}$ is the estimated probability for each class, computed as in Eq. (\ref{equ:prob}). $\lambda_r$ is the coefficient for $L_2$ regularization.

We use the backpropagation method to compute the gradients and update all the parameters $\Theta$ by:
\begin{equation}
\Theta = \Theta - \lambda_l  \frac{\partial J(\Theta)}{\partial\Theta}
\end{equation}
where $\lambda_l$ is the learning rate.

In order to avoid overfitting, we use dropout strategy to randomly omit half of the feature detectors on each training case. 
After learning  $\Theta$, we test the instance by feeding the target with its contexts into the IAN model, and the label with the highest probability stands for the predicted sentiment polarity of the target. 

\section{Experiments} 
\subsection{Experiment Preparation}
\subsubsection*{Dataset} 
We conduct experiments on SemEval 2014 Task 4\footnote{The detail introduction of this task can be seen at: http://alt.qcri.org/semeval2014/task4/} to validate the effectiveness of our model. 
The  SemEval 2014  dataset is composed of reviews in two categories: \emph{Restaurant} and \emph{Laptop}.
The reviews are labeled with three sentiment polarities: \emph{positive}, \emph{neutral} and \emph{negative}. 

\subsubsection*{Evaluation Metric} 
To evaluate the performance of aspect-level sentiment classification. we adopt the  \emph{Accuracy} metric, which is defined as:
\begin{equation}
Acc = \frac{T}{N}
\end{equation}
where $T$ is the number of correctly predicted samples, $N$ is the total number of samples.
\emph{Accuracy} measures the percentage of correct predicted samples in all samples. Generally, a well performed system has a higher accuracy.

Table \ref{tab:stat} shows the  training and test instance numbers in each category.
\begin{table}[htb]
	\centering
	\begin{tabular}{|l|l|l|l|l|l|l|}
		\hline
		\multirow{2}{*}{Dataset} & \multicolumn{2}{c|}{Positive} & \multicolumn{2}{c|}{Neural} & \multicolumn{2}{c|}{Negative} \\ \cline{2-7} 
		& \multicolumn{1}{l|}{Train}    & \multicolumn{1}{l|}{Test}   & \multicolumn{1}{l|}{Train} 
		& \multicolumn{1}{l|}{Test}     & \multicolumn{1}{l|}{Train}  & \multicolumn{1}{l|}{Test} \\ \hline
		Restaurant              &2164                           &728                          &637                                     
		&196                            &807                          &196                        \\ \hline
		Laptop                  &994                            &341                          &464                                    
		&169                            &870                          &128                        \\ \hline
	\end{tabular}
	\caption{Statistics of SemEval 2014 Dataset.}
	\label{tab:stat}
\end{table}

\subsubsection*{Hyperparameters Setting}
In our experiments, all word embeddings from context and target are initialized by GloVe\footnote{Pre-trained word embedding of GloVe can download from http://nlp.stanford.edu/projects/glove/}~\cite{pennington2014glove}, and all out-of-vocabulary words are initialized by  sampling from the uniform distribution $U(-0.1, 0.1)$. All weight matrices are given their initial values by sampling from uniform distribution $U(-0.1, 0.1)$, and all biases are set to zeros.
The dimensions of word embeddings, attention vectors and LSTM hidden states are set to 300 as in~\cite{wang2016attention}.
To train the parameters of IAN, we employ the Momentum~\cite{qian1999momentum}, which adds a fraction $\gamma$ of the update vector in the prior step to the current update vector.
The coefficient of $L_2$ normalization in the objective function is set to $10^{-5}$, and the dropout rate is set to 0.5.

\subsection{Model Comparisons}
In order to comprehensively evaluate the performance of IAN, we list some baseline approaches for comparison.
The baselines are introduced as follows.

$\bullet$ \textbf{Majority} is a basic baseline method, which assigns the largest sentiment polarity in the training set to each sample in the test set.

$\bullet$ \textbf{LSTM} only uses one LSTM network to model the context and get the hidden state of each word. After that, the average value of all the hidden states is regarded as final representation  and fed to a softmax function to estimate the probability of each sentiment label~\cite{wang2016attention}.

$\bullet$ \textbf{TD-LSTM} adopts two  long short-term memory (LSTM) networks to model the left context with target and the right context with target respectively.
The left and right target-dependent representations are concatenated for predicting the sentiment polarity of the target~\cite{tang2015effective}.

$\bullet$ \textbf{AE-LSTM} represents targets with aspect embeddings. First this method models the context words via LSTM networks and then combine the word hidden states with aspect embeddings to supervise the generation of attention vectors, which are in turn used to produce the final representation for aspect-level sentiment classification~\cite{wang2016attention}. 

$\bullet$ \textbf{ATAE-LSTM} is developed based on AE-LSTM.  ATAE-LSTM further strengthens the effect of aspect embeddings and  appends the aspect embeddings with each word embedding  vector to represent the context. The other design of ATAE-LSTM is the same as AE-LSTM~\cite{wang2016attention}.

\begin{table}[tbp]
	\begin{center}
    		\begin{tabular}{|l|c|c|}
			\hline
			Dataset           &Restaurant        &Laptop          \\ \hline \hline
			Majority          &0.535              &0.650   			 \\ \hline
			LSTM 	       	  &0.743              &0.665             \\ \hline
			TD-LSTM           &0.756	          &0.681             \\ \hline
			AE-LSTM           &0.762 	          &0.689             \\ \hline
			ATAE-LSTM         &0.772	          &0.687             \\ \hline
			IAN               &\textbf{0.786}     &\textbf{0.721}    \\ \hline
		\end{tabular}
		\caption{Comparison with baselines. Accuracy on 3-class prediction about SemEval 2014 Task 4 which includes restaurants and laptops. Best performances are in \textbf{bold}.}
		\label{tab:result}
	\end{center}
\end{table}

Table \ref{tab:result} shows the performance comparison of IAN with other baselines. 
From Table \ref{tab:result}, we can observe that, 
the Majority method is the worst, meaning the majority sentiment polarity occupies 53.5\% and 65.0\% of all samples in the Restaurant and Laptop categories respectively. All the other methods are  based on LSTM models and better than the Majority method, showing that LSTM has potentials in  automatically generating representations and can all bring performance improvement for sentiment classification.

The LSTM method gets the worst performance of all the neural network baseline methods, because it treats targets equally with other context words and does not make full use of the target information. This also verifies the work of ~\cite{jiang2011target} which points out the importance of targets.

TD-LSTM outperforms LSTM over 1 percent and 2 percent on the Restaurant and Laptop category respectively, since it develops from the standard LSTM and  processes the left and right contexts with targets. As we know, targets are twice represented and in some sense are specifically focused in the final representation.

Further, both AE-LSTM and ATAE-LSTM stably exceed the TD-LSTM method because of the introduction of attention mechanism.
For AE-LSTM and ATAE-LSTM, they capture important information in the context with the supervision of target and generate more reasonable representations for aspect-level sentiment classification.
We can also see that AE-LSTM and ATAE-LSTM further emphasize the modeling of targets via the addition of the aspect embedding, which is also the reason of performance improvement.
Compared with AE-LSTM, ATAE-LSTM especially enhance the interaction between the context words and target and thus has a better performance than AE-LSTM.

Our IAN model takes a further step towards emphasizing the importance of targets through learning target and context representation interactively.
We can see that IAN  achieves the best performance among all baselines. Compared with ATAE-LSTM model, IAN improves the performance about 1.4\% and 3.2\% on the \emph{Restaurant} and \emph{Laptop} categories respectively. 
As we know, it is difficult to boost 1 percent of accuracy on sentiment classification.
The main reason may be that IAN models the target and context using two connected attention networks which can influence each other.
With this design, we can well learn the representations of targets and contexts whose collocation contributes to aspect-level sentiment classification.
From Table  \ref{tab:result}, targets are progressively emphasized among these methods. The more attentions are paid to targets, the higher accuracy the system achieves.
This also inspires our future work to further research the modeling of targets.

\subsection{Analysis of IAN Model}

\begin{table}[tbp]
	\begin{center}
    		\begin{tabular}{|l|c|c|}
			\hline
			Dataset           &Restaurants        &Laptops           \\ \hline \hline
			No-Target         &0.772 	          &0.708             \\ \hline
			No-Interaction 	  &0.769              &0.706             \\ \hline
			Target2Content   &0.775	          &0.712             \\ \hline
			IAN               &\textbf{0.786}     &\textbf{0.721}    \\ \hline
		\end{tabular}
		\caption{Analysis of Interactive attention Networks.}
		\label{tab:analysis}
	\end{center}
\end{table}

\begin{table*}[]
\centering
\begin{tabular}{|l|l|l|l|l|l|l|}
\hline
\multicolumn{1}{|c|}{Dataset} & \multicolumn{1}{c|}{1} & \multicolumn{1}{c|}{2} & \multicolumn{1}{c|}{3} & \multicolumn{1}{c|}{4} & \multicolumn{1}{c|}{5} & \multicolumn{1}{c|}{\textgreater5} \\ \hline\hline
Restaurants-Train             & 2720/0.7539            & 604/0.1674             & 172/0.0477             & 56/0.0155              & 29/0.0080              & 27/0.0075                          \\ \hline
Restaurants-Test              & ~~801/0.7152             & 215/0.1920             & ~~57/0.0509              & 25/0.0223              & ~~8/0.0071               & 14/0.0125                          \\ \hline\hline
Laptops-Train                 & 1473/0.6327            & 649/0.2788             & 141/0.0606              & 52/0.0223              & ~~8/0.0035               & ~~5/0.0021                           \\ \hline
Laptops-Test                  & ~~351/0.5502             & 209/0.3276             & ~~45/0.0705              & 18/0.0282               & ~~9/0.0141               & ~~6/0.0094                           \\ \hline
\end{tabular}
\caption{Statistics of target length on SemEval 2014.}
\label{tab:target_length}
\end{table*}


In this section, we design a series of  models to verify the effectiveness of our IAN model. 
First, we ignore the modeling of targets and design a No-Target model that just uses the context representation.
Here we adopt only one LSTM networks with attention mechanism to model the context, where the attention vectors are computed through the averaged value of the target word embeddings. 
Then, we implement the second model No-Interaction which uses two LSTM networks to learn the representations of target and context via their own local attentions without any interaction.
Next, we design  Target2Content which also employs two LSTM networks to learn target and context representations, but only considers to attention context words via target representations.
The only difference between Target2Content and IAN is that IAN also adopts attention mechanism when modeling targets.
Table~\ref{tab:analysis} shows the performances of all these models.
From Table~\ref{tab:analysis}, we can see that the No-Target model achieves worse performances than the IAN model. The results verify that target  should be separately modeled and target representations can make contribution to judging the sentiment polarity of a target.

For the No-Interaction model, it gets the worst result among all the approaches. Compared with Target2Content and IAN, there is no interaction between target and context. Therefore, the interaction between target and content plays a great role in generating better representation for predicting target sentiment. Its results are also worse than No-Target, which may be because the effect of target representation, generated by local attention, is less important than the influence of target information for supervising learning content representation via attention.

As for Target2Content, it outperforms No-Interaction and is worse than IAN. Compare with IAN, it just learns the target representation via LSTM networks without supervision of context. As stated in Section 1, the collocated context and target can influence each other. 
That means that the interaction between target and content is critical to classifying target sentiment polarity and  unidirectional attentions are not enough to the final representation.

As we expect, IAN achieves the best performance among all the methods. This is because the IAN model fully considers the effect of the target and the interaction between target and context which makes contributions to the target sentiment classification.

Furthermore, from Table \ref{tab:result}, we can see that, the improvements on \emph{Restaurant} category is much less than those on \emph{Laptop} category. To explain this phenomenon, we analyze these two categories  and show the number and ratio of  targets  distributing in the two categories with respect to the target length (i.e., the word number included in a target), as in Table \ref{tab:target_length}. 
From Table \ref{tab:target_length}, we can see that, the number of instances with 1-word target in \emph{Restaurant} category is 9\% more than that in the \emph{Laptop} category. This means that \emph{Laptop} category has more multi-words targets than \emph{Restaurant} category. 
In IAN, we model the targets by LSTM networks and interactive attentions.
LSTM networks and interactive attention are more effective on modeling  long targets than short targets. 
Conversely, average/max pooling, used in other methods, usually lose more information in modeling long targets compared with shorter target. 
This demonstrates the effectiveness of separately modeling the targets via LSTM networks and interactive attention.

\subsection{Case Study}

\begin{figure*}[htb]
\centering
	\includegraphics[width=1.0\linewidth]{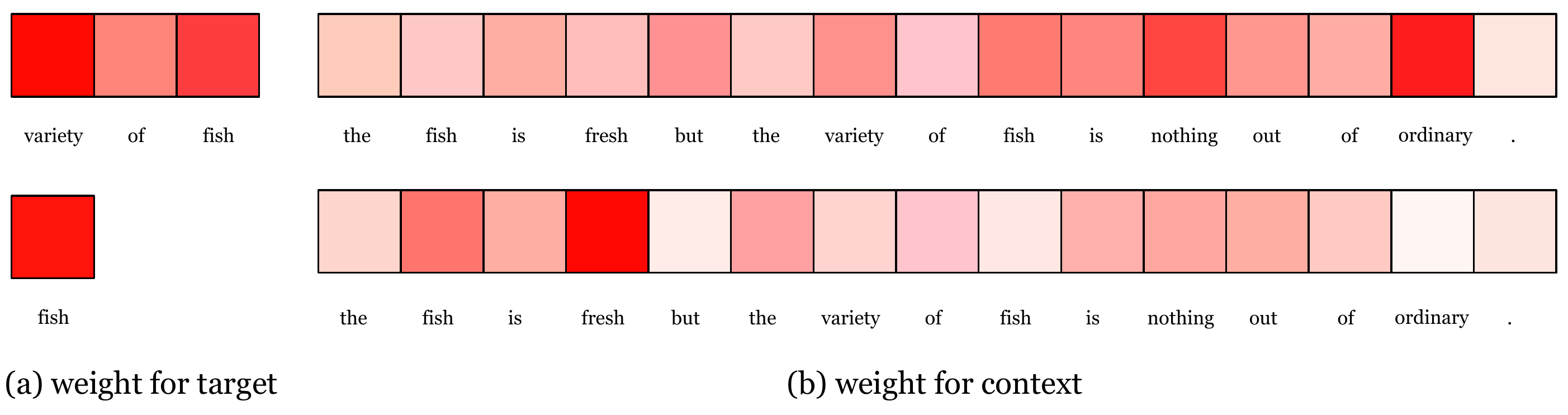}
\caption{Case Study: Illustration of Attention Weights for \emph{Context} and \emph{Target}.} 
\label{fig:case}
\end{figure*}
In this section, we use a review context ``\textbf{\emph{the fish is fresh but the variety of fish is noting out of ordinary.}}'' and two targets ``\textbf{\emph{fish}}'' and ``\textbf{\emph{variety of fish}}'' from Restaurant category as a case study. 
We apply IAN to model the \emph{context} and \emph{target}, and get the correct sentiment polarity: \emph{negative} and \emph{positive} for two targets respectively.
Figure \ref{fig:case} visualizes the attention weights on the \emph{context} and \emph{targets} computed by IAN. 
The left figure (a) gives the attention weights for two targets, and the right figure (b) shows the  attention weights of the context. Each line presents the corresponding target and context pairs.
The deeper color means the higher weight.

From Figure \ref{fig:case}, we can observe that the common words \emph{``the''} and \emph{``of''} and punctuation \emph{``.''} are paid little attention by IAN in the context. This verifies our intuition that some common words and punctuations makes little contributions to judging target sentiment polarity. 
The meaning of the context in the case study should be that the quality of \emph{fish} is good but the \emph{variety of fish} has nothing special.
 Obviously, the words \emph{fresh}, \emph{nothing, out, ordinary} play a great role in the sentiment classification of \emph{fish} and \emph{variety of fish}, and our model pays much attention on those words as we expect. 
 In addition, IAN also gives attention to the evaluated object: \emph{fish} and \emph{variety of fish}. Furthermore, when IAN is applied to the \emph{fish}, the ``attention'' of the model is mainly paid to the corresponding target and its real collocation, and little attention is given to the \emph{variety of fish} and its collocation. The situation is in contrast when  judging the target \emph{variety of fish}.

For the target \emph{variety of fish}, \emph{variety} is the head word, and other words are  used to modify the head word. Therefore, \emph{``variety''} is more important for expressing the target than the other two words  \emph{``of''} and \emph{``fish''}. From  Figure \ref{fig:case}, we can see that the IAN pays more attention to the word \emph{``variety''} than to the other words.
This means that in our model, a target can provide useful information for its context to tune its attentions, and the context also plays an important role in supervising a target to get its focus.
Thus, through the IAN, we can well model  targets and contexts separately and interactively, and the concatenated representation of target and context are helpful for the aspect-level  sentiment classification.

\section{Related Work}
Aspect-level sentiment classification is a branch of sentiment analysis, and its research approaches can be split into two directions: traditional machine learning methods and neural networks methods. 

Aspect-level sentiment classification is typically regarded as text classification problem. Accordingly, text classification methods, such as SVM~\cite{pang2002thumbs}, can be applied to solve the aspect-level sentiment classification task without consideration of the mentioned target or aspect.
Traditional machine learning methods mainly focus on extracting a set of features like sentiment lexicons features and bag-of-word features to train a sentiment classifier~\cite{rao2009semi,kaji2007building,jiang2011target,perez2012learning,mohammad2013nrc}.
 Although these methods achieve a comparable performance, their results highly depend on the effectiveness of the handcraft features which are labor intensive. 

Recently, kinds of neural networks methods, such as Recursive Neural networks~\cite{socher2011semi,dong2014adaptive,qian2015learning}, Recursive Neural Tensor Networks~\cite{socher2013recursive}, Recurrent Neural networks~\cite{mikolov2010recurrent,tang2015effective}, Tree-LSTMs~\cite{tai2015improved} and Hierarchical LSTMs~\cite{ruder2016hierarchical}, have achieved a promising result on sentiment analysis. However, the neural network based  approaches just make use of the contexts without consideration of  targets which also make great contributions to judging the sentiment polarity of target. 

To the best of our knowledge, Jiang et al. \shortcite{jiang2011target} first present the importance of targets in sentiment classification and argue that 40\% of sentiment classification errors are caused by not considering targets. 
Later, in order to incorporate  targets into a model, Tang et al.~\shortcite{tang2015effective} develop two target-dependent long short-term memory (LSTM) to model the left and right contexts with target, where the target information is automatically taken into account.
In addition, Tang et al.~\shortcite{tang2016aspect} designs deep memory networks which consist of multiple computational layers to integrate the target information. Each layer is a context- and location- based attention model, which first learns the importance/weight of each context word and then utilizes the information to calculate  context representation.
To take target into account, Wang et al.~\shortcite{wang2016attention} explore the potential correlation of targets and sentiment polarities in aspect-level sentiment classification. 
In order to capture important information in response to a given target, they design an attention-based LSTM to concentrate on different parts of a sentence when different targets are taken as input.

As described in Section~\ref{section:intro}, target can help to attention the closely related part in the context. Therefore, we build an interactive attention network (IAN) model which respectively utilizes the target and context to compute the attention vector and learn the target and context representations. In this way, IAN can well acquire the appropriate final representations for aspect-level sentiment classification compared with other methods.

\section{Conclusion}
In this paper, we design an interactive attention networks~(IAN) model for aspect-level sentiment classification.
The main idea of IAN is to use two attention networks to model the target and context interactively. 
The IAN model can pay close attention to the important parts in the  target and context and well generate the representations of target and context.
Then, IAN benefits from the target representation which is always ignored in other methods.
Experiments on SemEval 2014 verify that IAN can learn effective features for target and content and provide enough information for judging the target sentiment polarity.
The case study also shows that IAN  can reasonably pay attention to those words which are important to judging the sentiment polarity of targets. 

 \section*{Acknowledgements}
 Our work is supported by Major National Social Science Fund of China~(No.12\&ZD227) and National Natural Science Foundation of China~(No.61370117 \& No.61572049). The corresponding authors of this paper are Houfeng Wang \& Sujian Li.
\bibliographystyle{named}
\bibliography{ijcai17}

\end{document}